# Towards Natural Language Question Answering Over Earth Observation Linked Data Using Attention-Based Neural Machine Translation

*Abhishek V. Potnis, Rajat C. Shinde, Surya S. Durbha*

Centre of Studies in Resources Engineering, Indian Institute of Technology Bombay, India
{abhishekvpotnis, rajatshinde, sdurbha}@iitb.ac.in

## ABSTRACT

With an increase in Geospatial Linked Open Data being adopted and published over the web, there is a need to develop intuitive interfaces and systems for seamless and efficient exploratory analysis of such rich heterogeneous multi-modal datasets. This work is geared towards improving the exploration process of Earth Observation (EO) Linked Data by developing a natural language interface to facilitate querying. Questions asked over Earth Observation Linked Data have an inherent spatio-temporal dimension and can be represented using GeoSPARQL. This paper seeks to study and analyze the use of RNN-based neural machine translation with attention for transforming natural language questions into GeoSPARQL queries. Specifically, it aims to assess the feasibility of a neural approach for identifying and mapping spatial predicates in natural language to GeoSPARQL's topology vocabulary extension including - Egenhofer and RCC8 relations. The queries can then be executed over a triple store to yield answers for the natural language questions. A dataset consisting of mappings from natural language questions to GeoSPARQL queries over the Corine Land Cover(CLC) Linked Data has been created to train and validate the deep neural network. From our experiments, it is evident that neural machine translation with attention is a promising approach for the task of translating spatial predicates in natural language questions to GeoSPARQL queries.

*Index Terms*— natural language, question-answering, earth observation, linked data

## 1. INTRODUCTION

There have been significant efforts in the research community as well as a part of various Governance initiatives[1], to encourage publishing of data as Linked Open Data. The value addition with Linked Open Data in terms of seamless integration, data interoperability, distribution and innovative application development has been well understood. Geospatial Earth Observation(EO)

```
Natural Language Question: Which are the areas that have
mixed forests adjacent to mineral extraction sites?

GeoSPARQL Query:

select distinct ?area1 ?area2
where {
?area1  corine:hasLandUse  corine:MixedForest .
?area2  corine:hasLandUse  corine:MineralExtractionSites .
?area1  corine:hasGeometry  ?geom1 .
?area2  corine:hasGeometry  ?geom2 .
filter (geof:sfTouches( ?geom1, ?geom2)) }
```

Fig. 1. Example Natural Language Question and its Equivalent GeoSPARQL Query with RCC8 Predicate over the Corine Land Cover(CLC) Linked Dataset

Linked Data in particular possess tremendous value due to its heterogeneity and multimodality. With the increasing size of the Geospatial Linked Open Data Cloud, there is a need to design and develop intuitive interfaces and systems for effective data exploration and analysis.

With advancements in Computer Vision, the area of Natural Language Visual Question Answering(NL-VQA) has received a lot of attention from the scientific community. A natural language interface for data exploration has been known to shield users from technical jargons and enable them to effectively utilize the underlying knowledge for their applications. Although there exist significant contributions to NL-QA over multimedia images, the research area of NL-QA over geospatial data is still in its nascent stage. Lobry et. al. in [2], propose a Convolutional Neural Network(CNN)-Recurrent Neural Network(RNN) based approach for feature extraction and mapping a natural language question to an answer over a remote sensing image.

This work focuses on Question Answering over Geospatial EO Linked Data. Generally, Linked Datasets are housed in databases - termed as Triple Stores. Triple Stores are graph databases that store knowledge in the form of triples $<s,p,o>$. Question Answering over Relational Databases has also seen tremendous progress, particularly in recent years, due to the popularization of Artificial Neural Networks.

Recurrent Neural Networks in particular have known to work extremely well with sequences of data. The Seq2Seq[3] model forms the basis of the state of the art methods in translating natural language questions to SQL queries. The Seq2SQL[4] neural network trained and validated over the WikiSQL dataset proposes to improve SQL queries generation from natural language questions using the reward based Reinforcement Learning technique.

In the area of Question Answering over Linked Data, Semantic Parsing - the process of transforming natural language into a machine readable logical form has been applied[5] for generating SPARQL(SPARQL Protocol and RDF Query Language) queries from natural language. There have also been efforts to use Neural Machine Translation for SPARQL Query Constructions[6], by treating SPARQL as a foriegn language[7] for translation. Although neural machine translation for SPARQL has been understood, its geospatial sibling - GeoSPARQL with spatial predicates for Geospatial Linked Data, remains largely unexplored.

Our contributions in this work are two-fold - (1) We examine the feasibility of attention-based neural machine translation for identification and mapping of spatial predicates in natural language to GeoSPARQL's topology vocabulary extension including Egenhofer and RCC8 relations. (2) We create a dataset consisting of natural language questions with their GeoSPARQL query equivalents for training and validation of the deep neural network. We report and discuss our findings from the experiments with attention based neural machine translation over the Corine Land Cover(CLC) Linked Dataset.

## 2. METHODOLOGY

### 2.1. Neural Machine Translation with Attention

Neural Machine Translation Networks with Attention have achieved state of the art performance for translating text between different languages.

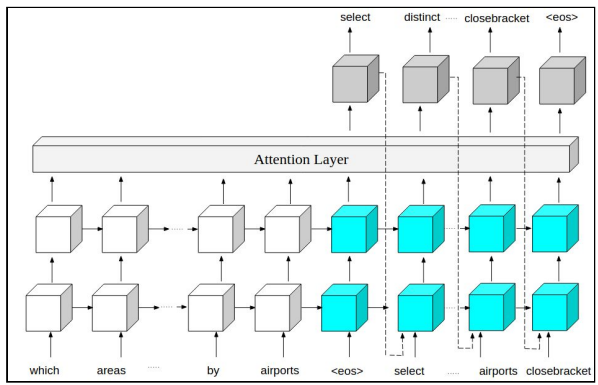

Fig. 2. Neural Machine Translation with Attention - Stacking of RNNs as an Encoder-Decoder Architecture to translate source Natural Language sequence to target GeoSPARQL sequence

Although the encoder-decoder based RNN architecture has proved to be effective for machine translation, it has been observed that their performance deteriorates with longer input sequences[8]. To address this issue, [9] and [10] propose attention based NMT that enables the model to allot importance to specific words in the input sequence as each word in the output sequence is being predicted. The importance allotted to words in the input sequence is quantified by assigning weights to them - termed as *attention weights*. These *attention weights* are then used by the decoder to predict words in the output sequence.

Figure 2 depicts the NMT with attention architecture for translating natural language questions to GeoSPARQL queries. The bidirectional recurrent neural network encoder-decoder architecture with Bahdanau Attention mechanism has been implemented for this study. The network has been trained for 200 epochs over the CLC Linked Data based dataset consisting of natural language questions and GeoSPARQL queries. Sparse Categorical Cross Entropy has been used as the Loss Function with Adam Optimizer.

## 3. EXPERIMENTAL RESULTS

### 3.1. Dataset

The Corine Land Cover (CLC)[1] Linked Geospatial Data published under the European FP7 project TELEIOS has been used for this study. The dataset consists of 44 land cover classes spanning a three-level nomenclature hierarchy.

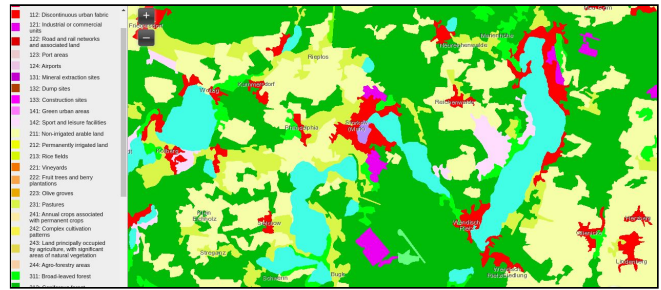

Fig. 3. Map View[2] of the Corine Land Cover (CLC) Dataset

The CLC Geospatial Linked Data conforming to the Corine Ontology[3] consist of Resource Description Framework(RDF) triples enumerating different land cover classes in the form of areas. Figure 3 is the map view of the CLC dataset representing different land cover classes. Figure 4 depicts the RDF snippet from the CLC Linked Data. It refers to an Area with an ID as "Area_0", which has a Polygon Geometry associated with it, and has its Land Use as "Continuous Urban Fabric". Geospatial Linked Data can be queried upon using GeoSPARQL. GeoSPARQL is a structured query language belonging to the SPARQL family, with special emphasis on handling geospatial data.

---

[1] https://old.datahub.io/dataset/corine-land-cover
[2] https://land.copernicus.eu/pan-european/corine-land-cover/clc2018
[3] http://pyravlos-vm5.di.uoa.gr/corineLandCover.svg

```
<http://geo.linkedopendata.gr/corine/Area_0>
<http://geo.linkedopendata.gr/corine/ontology#hasGeometry> "POLYGON
((25.124978607257269 35.335507039952923,25.125731428234683
… 35.33657416799705,35.124245480098633,
35.336224035643134,25.124978607257269))"
^^<http://strdf.di.uoa.gr/ontology#WKT> .
<http://geo.linkedopendata.gr/corine/Area_0>
<http://geo.linkedopendata.gr/corine/ontology#hasLandUse>
<http://geo.linkedopendata.gr/corine/ontology#continuousUrbanFabric>
```

Fig. 4. Snippet of the CLC Geospatial Linked Data

### 3.1.1. GeoSPARQL Encoding

With an objective to create a dataset consisting of mappings from natural language questions to equivalent GeoSPAQL queries over the CLC Linked data, GeoSPARQL encoding has been performed to normalize the text in the query. Encoding GeoSPARQL to normalized text has been known to aid in the tokenization process prior to training of the neural network model. Figure 5 is an example of an encoded GeoSPARQL query with its original query. The encoding involves replacement of non alphabetic characters such as question mark, curly brackets, colon and parentheses with appropriate alphabetical words.

```
GeoSPARQL Query:

select distinct ?area1 ?area2
where {
?area1  corine:hasLandUse  corine:MixedForest .
?area2  corine:hasLandUse  corine:MineralExtractionSites .
?area1  corine:hasGeometry  ?geom1 .
?area2  corine:hasGeometry  ?geom2 .
filter (geof:sfTouches( ?geom1, ?geom2)) }

Encoded GeoSPARQL Query:

select distinct varAreaOne varAreaTwo
where openBracket
varAreaOne corine hasLandUse corine MixedForest dot
varAreaTwo corine hasLandUse corine MineralExtractionSites
dot varAreaOne corine hasGeometry varGeomOne dot
varAreaTwo corine hasGeometry varGeomTwo dot
filter openParanthesis geof sfTouches openParanthesis
varGeomOne comma varGeomTwo closeParanthesis
closeParanthesis closeBracket
```

Fig. 5. Example of GeoSPARQL Encoding to aid in the Tokenization for Neural Machine Translation

### 3.1.2. Dataset Development

The dataset consists of 528 natural language to GeoSPARQL query pairs, each of which have been manually created and validated against the CLC Linked Dataset. The dataset covers the 3 'Wh' - 'What', 'Where' and 'Which' questions including up to 5 paraphrases of each question. This dataset although not comprehensive, is envisaged as a preliminary step towards improving question answering over EO Linked Data.

```
Natural Language Question:
which areas are covered by Airports

GeoSPARQL Query:
select distinct varArea where
openBracket
varArea corine hasLandUse corine Airports
closeBracket
```

Fig. 6. Example of Natural Language Question and its Equivalent GeoSPARQL Query from the dataset

Around 60% of the queries in the dataset use GeoSPARQL spatial predicates - *'geof:sfContains'* or *'geof:sfTouches'* for mapping natural language questions of spatial nature. The 80-20 split ratio has been used to randomly split the dataset for training and validation.

## 3.2. Evaluation and Discussion

### 3.2.1. Bilingual Evaluation Understudy Score

The model has been evaluated over our data set for computing the Bilingual Evaluation Understudy(BLEU) score. Table 1 depicts the individual and cumulative BLEU scores, with the overall BLEU score of 0.8179 being achieved over the validation dataset.

Table 1: BLEU Score of the Validation Dataset using NMT with Attention

| BLEU | 81.79 | | | |
|---|---|---|---|---|
| Type | 1-gram | 2-gram | 3-gram | 4-gram |
| Individual | 83.74 | 82.01 | 81.14 | 80.29 |
| Cumulative | 83.74 | 82.87 | 82.29 | 81.79 |

Figure 7 represents a natural language question posed as the input to the trained model and its predicted GeoSPARQL query. It is interesting to note that the model correctly predicted the *geof:sfTouches* predicate equivalent to the RCC 8 relation *EC* of GeoSPARQL. Figure 8 depicts the visualization of attention of our model for the input natural language question and the predicted GeoSPARQL query. From the visualization, it is noteworthy that the model learned to translate the word 'adjacent' in natural language to the *geof:sfTouches* predicate of GeoSPARQL.

**Input Natural Language Question:**
what are the areas that have constructionsites adjacent to mixedforest <end>

**Predicted GeoSPARQL Query:**
select distinct varareaone varareatwo
where openbracket
varareaone corine haslanduse corine constructionsites dot
varareatwo corine haslanduse corine mixedforest dot
varareaone corine hasgeometry vargeomone dot
varareatwo corine hasgeometry vargeomtwo dot
filter openparanthesis geof sftouches openparanthesis
vargeomone comma vargeomtwo closeparanthesis
closeparanthesis closebracket <end>

Fig. 7. Input Natural Language Question and its Predicted GeoSPARQL Query with Spatial Predicate

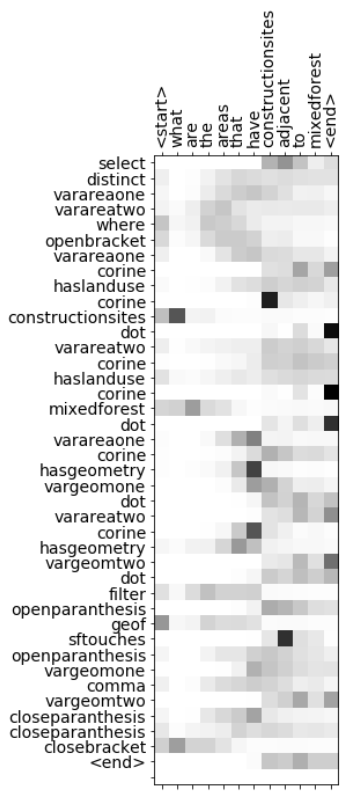

Fig. 8. Attention Visualization of Input NL Question and its Predicted GeoSPARQL Query Translation with Spatial Predicate

## 4. CONCLUSION

In the background of increasing adoption of Geospatial Linked Open Data, this work advocates the development of intuitive and seamless systems for effective exploration of such heterogeneous and multimodal datasets. In that regard, this work proposes to develop a natural language interface for querying over geospatial linked data. Although Geospatial Linked Data can be queried upon using GeoSPARQL, a background of the query language in addition to knowledge and understanding of databases is essential for this purpose. Thus a natural language interface would shield a prospective user from the technical know-how and also enable her/him to effectively utilize the underlying knowledge residing in the database. The paper discusses the neural machine translation with attention approach for translating natural language questions to GeoSPARQL queries, specifically focussing on mapping spatial predicates of natural language to Egenhofer and RCC8 predicates of GeoSPARQL. From the experiments with our dataset, NMT with attention seems a promising approach for natural language interfacing with GeoSPARQL. The dataset created as a part of this work is envisaged to be enriched further to improve the diversity of natural language questions and contribute to the area of natural language question answering over EO Linked Data.

## 5. ACKNOWLEDGMENTS

The authors express their gratitude to the Management of Data, Information and Knowledge Group (MaDgIK) of the Department of Informatics and Telecommunications of the National and Kapodistrian University of Athens for creating and publishing the Corine Land Cover(CLC) Linked Dataset.

## 6. REFERENCES

[1] F. Kirstein, B. Dittwald, S. Dutkowski, Y. Glikman, S. Schimmler, and M. Hauswirth, "Linked Data in the European Data Portal: A Comprehensive Platform for Applying DCAT-AP," *International Conference on Electronic Government*, 2019.

[2] S. Lobry, J. Murray, D. Marcos and D. Tuia, "Visual Question Answering From Remote Sensing Images," *IGARSS 2019 - 2019 IEEE International Geoscience and Remote Sensing Symposium*, Yokohama, Japan, pp. 4951-4954, 2019.

[3] S. Ilya, B. Oriol, Q. VV Le, "Sequence to sequence learning with neural networks," *Advances in Neural Information Processing Systems*, pp. 3104-3112, 2014.

[4] Z. Victor, C. Xiong, and R. Socher. "Seq2sql: Generating structured queries from natural language using reinforcement learning." *arXiv preprint arXiv:1709.00103*, 2017.

[5] F. F. Luz, M. Finger, "Semantic parsing natural language into sparql: improving target language representation with neural attention". *arXiv preprint arXiv:1803.04329*, 2018.

[6] T. Soru, E. Marx, , A. Valdestilhas, , D. Esteves, , D. Moussallem, G. Publio,"Neural Machine Translation for Query Construction and Composition". *arXiv preprint arXiv:1806.10478*, 2018

[7] T. Soru, E. Marx, D. Moussallem, G. Publio, A. Valdestilhas, , D. Esteves, C. B. Neto, "SPARQL as a Foreign Language". *arXiv preprint arXiv:1708.07624*, 2017

[8] K. Cho, B. van Merrienboer, D. Bahdanau, Y. Bengio, "On the properties of neural machine translation: Encoder-decoder approaches", *arXiv preprint arXiv:1409.1259*, 2014

[9] D. Bahdanau, K. Cho, Y. Bengio, "Neural Machine Translation by Jointly Learning to Align and Translate", *International Conference on Learning Representations*, 2015.

[10] T. Luong, H. Pham, C. D. Manning, "Effective approaches to attention-based neural machine translation", *Proc. Empirical Methods Natural Lang. Process.*, pp. 1412-1421, 2015.